\documentclass[a4paper,twoside]{article}

\usepackage{bera}
\usepackage{listings}
\usepackage{xcolor}
\colorlet{punct}{red!60!black}
\definecolor{background}{HTML}{EEEEEE}
\definecolor{delim}{RGB}{20,105,176}
\colorlet{numb}{magenta!60!black}
\usepackage{float}

\usepackage{epsfig}
\usepackage{subcaption}
\usepackage{calc}
\usepackage{amssymb}
\usepackage{amstext}
\usepackage{amsmath}
\usepackage{amsthm}
\usepackage{multicol}
\usepackage{pslatex}
\usepackage{apalike}
\usepackage[bottom]{footmisc}
\usepackage{SCITEPRESS}     

\lstdefinelanguage{json}{
    basicstyle=\normalfont\ttfamily,
    numbers=left,
    numberstyle=\scriptsize,
    stepnumber=1,
    numbersep=0.3pt,
    showstringspaces=false,
    breaklines=true,
    frame=lines,
    literate=
     *{0}{{{\color{numb}0}}}{1}
      {1}{{{\color{numb}1}}}{1}
      {2}{{{\color{numb}2}}}{1}
      {3}{{{\color{numb}3}}}{1}
      {4}{{{\color{numb}4}}}{1}
      {5}{{{\color{numb}5}}}{1}
      {6}{{{\color{numb}6}}}{1}
      {7}{{{\color{numb}7}}}{1}
      {8}{{{\color{numb}8}}}{1}
      {9}{{{\color{numb}9}}}{1}
      {:}{{{\color{punct}{:}}}}{1}
      {,}{{{\color{punct}{,}}}}{1}
      {\{}{{{\color{delim}{\{}}}}{1}
      {\}}{{{\color{delim}{\}}}}}{1}
      {[}{{{\color{delim}{[}}}}{1}
   }

\begin{document}

\title{Towards view-invariant vehicle speed detection from driving simulator images}

\author{\authorname{Antonio Hernández Martínez\sup{1}\orcidAuthor{0000-0001-5722-1531}, David Fernández Llorca \sup{1, 2}\orcidAuthor{ 0000-0003-2433-7110} and Iván García Daza\sup{1}\orcidAuthor{0000-0001-8940-6434}}
\affiliation{\sup{1}Computer Engineering Department, University of Alcalá, Alcalá de Henares, Spain}
\affiliation{\sup{2}European Commission, Joint Research Centre (JRC), Seville}
\email{\{antonio.hernandezm, david.fernandezl, ivan.garciad\}@uah.es}
}


\abstract{The use of cameras for vehicle speed measurement is much more cost effective compared to other technologies such as inductive loops, radar or laser. However, accurate speed measurement remains a challenge due to the inherent limitations of cameras to provide accurate range estimates. In addition, classical vision-based methods are very sensitive to extrinsic calibration between the camera and the road. In this context, the use of data-driven approaches appears as an interesting alternative. However, data collection requires a complex and costly setup to record videos under real traffic conditions from the camera synchronized with a high-precision speed sensor to generate the ground truth speed values. It has recently been demonstrated \cite{Hernandez2021} that the use of driving simulators (e.g., CARLA) can serve as a robust alternative for generating large synthetic datasets to enable the application of deep learning techniques for vehicle speed estimation for a single camera. In this paper, we study the same problem using multiple cameras in different virtual locations and with different extrinsic parameters. We address the question of whether complex 3D-CNN architectures are capable of implicitly learning view-invariant speeds using a single model, or whether view-specific models are more appropriate. The results are very promising as they show that a single model with data from multiple views reports even better accuracy than camera-specific models, paving the way towards a view-invariant vehicle speed measurement system.  }

\onecolumn \maketitle \normalsize \setcounter{footnote}{0} \vfill

\section{\uppercase{Introduction}}
\label{sec:introduction}

A clear causal relationship between excess speed and the risk of death and serious injury is well-established. Thus, speed enforcement through automatic speed measurement, vehicle identification and subsequent sanctioning is a fundamental road safety measure that leads directly to the reduction of accidents \cite{ERSO2018}. In fact, in the vicinity of speed cameras, the reduction of speeding vehicles and collisions can reach 35\% and 25\% respectively \cite{Wilson2010}. In addition, it has been shown that the higher the intensity of the enforcement, the greater the reduction in accidents \cite{Elvik2011}. 

\begin{figure*}[ht]
  \centering
  \includegraphics[width=0.67\linewidth]{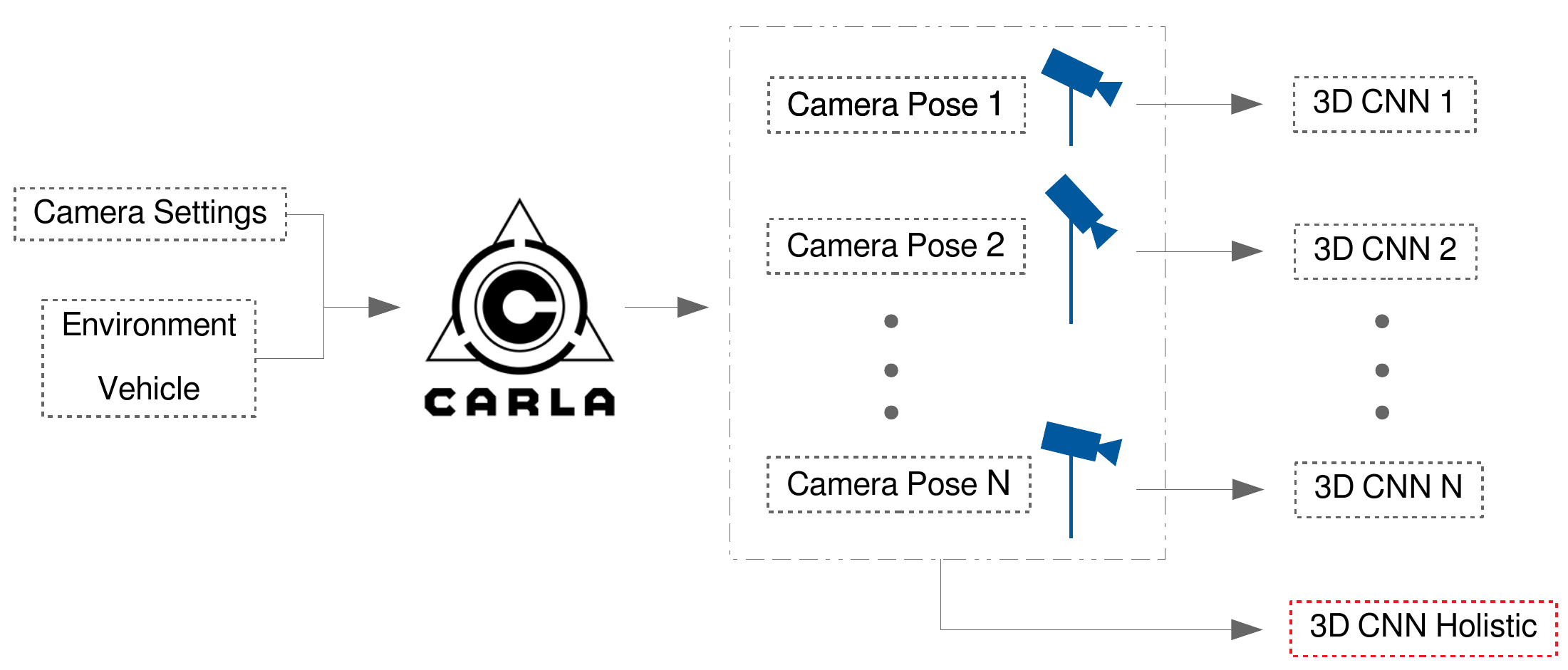}
  \caption{Overview of the presented approach. From left to right: control parameters, simulator, camera-specific datasets, view-specific models, and view invariant (holistic) model.}
  \label{fig:overview}
\end{figure*}

Most common speed measurement technologies include high-accuracy, high-cost range sensors such as radar and laser, as well as intrusive sensors such as magnetic inductive loop detectors. The use of cameras has rarely been proposed for speed detection due to the demanding requirements for accuracy and robustness, and the intrinsic limitations of cameras to accurately measure range. They often play a secondary role in detecting license plates or capturing human-understandable information from the scene, allowing visual identification of vehicles, even drivers, and serving as evidence when the speed limit is violated. Recent advances in computer vision, however, have led to a significant increase in the number of proposals that use vision as the only mechanism for measuring vehicle speed \cite{Llorca2021}. Cameras are a very cost-effective alternative that allows the integrated implementation of speed detection along with other functionalities such as license plate detection, automatic vehicle identification or even occupant detection.\newpage

Still, vision-based speed measurement is a challenging problem due to the discrete nature of video sensors. In addition, classical vision-based methods are very sensitive to extrinsic calibration between the camera and the road. In this context, the use of data-driven approaches appears as an interesting alternative but real data collection requires a complex setup \cite{Llorca2016}. It has recently been proved that the use of driving simulators (e.g., CARLA) can serve as a robust alternative for generating large synthetic datasets to enable the application of deep learning techniques for vehicle speed estimation from a single camera \cite{Hernandez2021} or from multiple views \cite{Barros2021}.

In this paper, we extend our previous work \cite{Hernandez2021} by addressing the question of whether complex end-to-end 3D-CNN architectures (i.e., regression problem, from sequences to speeds) are capable of implicitly learning view-invariant speeds using a single holistic model, or whether view-specific models are more appropriate. We extend our synthetic dataset based on CARLA simulator \cite{carla2017} using six different virtual cameras with variations in height and pitch angle. Then six view-specific models are developed and compared with one single multi-view model trained with sequences from all the cameras. An overall view of the proposed approach is depicted in Fig. \ref{fig:overview}. The results presented are very promising as they suggest that the model is capable not only of absorbing the variability of multiple views, but also of benefiting from it and from the increased availability of data, which opens the door to future view-invariant systems.

\section{\uppercase{Related Work}}
As the survey presented in \cite{Llorca2021} concludes, although we can find hundreds of works focused on vision-based vehicle speed detection, the available solutions are not sufficiently mature. A considerable number of papers present sub-optimal camera pose and settings resulting in very high meter-to-pixel ratios that are unlikely to provide accurate estimates of depth and speed. Performance and robustness in difficult lighting and weather conditions are often neglected. The sensitivity of each approach to pose and camera settings is not sufficiently addressed. 

In addition, the number of data-driven approaches, well consolidated in other areas of computer vision, is still very limited for speed regression. This may also be due to the lack of well-established datasets to train and compare the performance of different methods. In this section, we focus on recent learning-based methods for vision-based vehicle speed detection from the infrastructure, including available datasets. We refer to \cite{Llorca2021} for a more complete overview of the state-of-the-art.

\subsection{Learning-based approaches}
Regarding the estimation of vehicle speed using data-driven methods, only a few papers are found in the available literature. In \cite{Dong2019}, average traffic speed estimation is addressed as a video action recognition problem using 3D CNNs. They concatenate RGB and optical flow images and found that optical flow was more useful. They emphasize that the main limitation is overfitting due to the lack of available data. In \cite{Madhan2020} a Modular Neural Network (MNN) architecture is presented to perform joint vehicle type classification and speed detection. These two approaches are more suitable for average traffic speed detection, since the input data correspond to multiple lanes and vehicles. Some kind of mechanism is needed to generate regions of interests with sufficient spatio-temporal information \cite{Biparva2022} to distinguish different speeds for different types of vehicles and scenarios. This issue can be mitigated by using a camera setting with a lower meter-pixel ratio focusing on a single lane.

Recent works have proposed the use of LSTM-type recurrent architectures \cite{Parimi2021}, \cite{Hernandez2021}, but as demonstrated in \cite{Hernandez2021} with synthetic data, they seem to perform worse than other non-recurrent methods such as 3D CNNs. In \cite{Revaud2021} camera calibration, scene geometry and traffic speed detection are performed using a transformer network in a learning process from synthetic training data, considering that cars have similar and known 3D shapes with normalized dimensions. In \cite{RR2022}, after vehicle detection and tracking using YOLOv3 and Kalman filter, respectively, a linear regression model is used to estimate the vehicle speed. Other statistical and machine learning-based methods are also compared. Vehicle detection and tracking is performed in \cite{Barros2021} using Faster R-CNN and DeepSORT, respectively. Next, dense optical flow is extracted using FlowNet2, and finally, a modified VGG-16 deep network is used to perform speed regression.

\subsection{Datasets for speed detection}
So far, only two datasets with real sequences and ground truth values are publicly available. First, the \emph{BrnoCompSpeed} dataset \cite{Sochor2017}, which contains 21 sequences ($\sim$1 hour per sequence) with 1920 $\times$ 1080 pixels resolution images at 50 fps in highway scenarios. The speed ground truth is obtained using a laser-based light barrier system. Second, the \emph{UTFPR} dataset \cite{Luvizon2017}, which includes 5 hours of sequences with 1920 $\times$ 1080 pixels resolution images at 30 fps in an urban scenario. The ground truth speeds are obtained using inductive loop detectors. 

Therefore, as described above, the use of synthetic datasets is becoming increasingly prevalent for this problem. For example, in \cite{Lee2019} a CNN model to estimate the average speed of traffic from top-view images is trained using synthesized images, which are generated using a cycle-consistent adversarial network (Cycle-GAN). 
Synthetic scenes with a resolution of 1024 $\times$ 768 pixels covering multiple lanes with vehicles randomly placed on the road are used in \cite{Revaud2021} to train and test the method used to jointly deal with camera calibration and speed detection. In our previous work \cite{Hernandez2021}, a publicly available synthetic dataset was generated using CARLA simulator, using one fixed camera at 80 FPS with Full HD format (1920 $\times$1080), with variability corresponding to multiple speeds, different vehicle types and colors, and lighting and weather conditions. A similar approach was presented in \cite{Barros2021} including multiple cameras and generating more than 12K instances of vehicles speeds. Preliminary, it is observed that multi-camera variability does not negatively affect the results. However, this effect has not been sufficiently studied. 

\begin{figure}[H]
    \centering
    \includegraphics[width=1.0\linewidth]{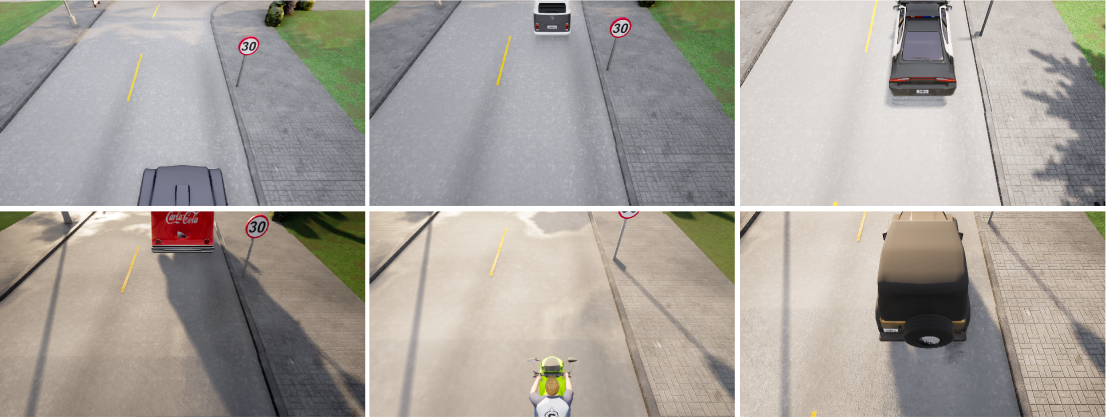}
    \caption{Overview of the different datasets. Upper row 4m camera height. Lower row 3m camera height. The columns show an 45º, 50º and 60º camera pitch from left to right.}
    \label{fig:CamPositions}
\end{figure}

\section{\uppercase{Method}}
 \label{sec:method}
This section describes the methodology used for the construction of the new dataset, based on the one shown in the previous work \cite{Hernandez2021}, including new camera poses, and increasing the complexity of the speed detection problem.

\subsection{Multi-view synthetic dataset}

Starting from the single-view synthetic dataset generated in the our previous work \cite{Hernandez2021} using the CARLA simulator \cite{carla2017}, the extrinsic parameters of the camera are modified, specifically the pitch angle and the height with respect to the road, leaving the rest of the extrinsic and intrinsic parameters fixed. 

\begin{figure}[t]
    \centering
    \includegraphics[width=0.7\linewidth]{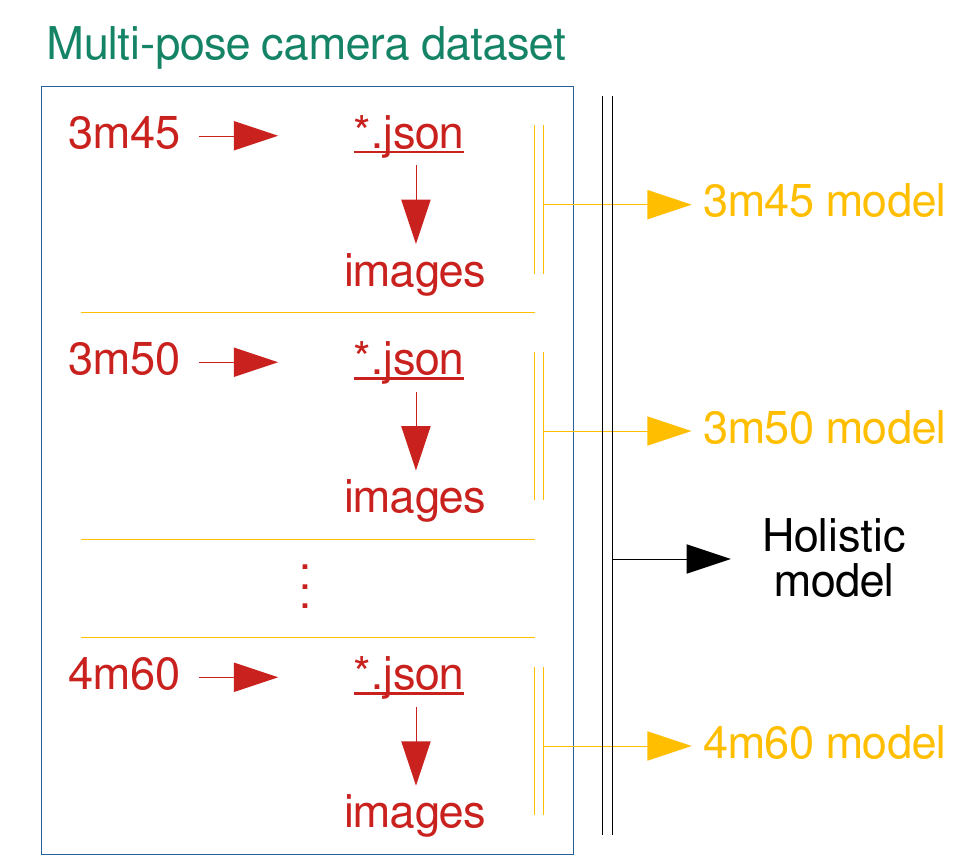}
    \caption{Schematic view of the dataset processing method.}
    \label{fig:scheme}
\end{figure}

As a first step to increase variability in a controlled manner and to be able to carry out a comprehensive analysis of results, we included six different poses, using two different camera heights and three different pitch angles (extrinsic parameters): 3m45 (the original one), 3m50, 3m60, 4m45, 4m50 and 4m60. On the one hand, a holistic model is generated, trained, validated and tested using all the sequences in the dataset. On the other hand, pose-specific models are trained, validated and tested only with the sequences corresponding to each pose. Fig. \ref{fig:scheme} depicts how the dataset is used.

For the selected camera heights (3m and 4m) the three pitch angels have been defined taking into account the road section captured by the camera field of view. For pitch angles greater than 60$^{\circ}$ the road section is too short, while for angles less than 45$^{\circ}$ the section is too long (for more details on the optimal distances for speed detection see \cite{Llorca2016}). With these configurations, road sections of between 4 and 10 meters in length were obtained. 

Therefore, six camera poses were used, generating 610 sequences per pose (total number of sequences is 3660). Each sequence with the associated vehicle has a random speed between 30 and 100 km/h. The sequence has an individual number of images depending on the speed. Thus, the number of images recorded in the dataset is approximately 427k. An example image is shown in Fig. \ref{fig:CamPositions} for each camera pose.

Each simulation carried out to obtain the image sequences generates a \emph{json} file. The \emph{json} files record the frame number, the brand and model of the simulated vehicle, as well as its speed and ID, among other keys. With the \emph{json} file, and using the ID, the labelling can be divided into sequences, each with the number of frames generated depending on the vehicle speed.

To generate the labelling of the holistic model, the \emph{json} files of each of the simulations are combined, adding for each sequence a new key indicating the dataset from which the sequence comes. In this way, each model can be evaluated individually, and cross tests can also be performed, as in the case of the holistic model, which has been used to test each of the datasets.

\begin{lstlisting}[caption={\emph{json} file labeling example.},label={fig:JsonEx},language=json,firstnumber=1]
{"elapsed seconds":2.05,
 "delta_seconds":0.0125,
 "platform_timestamp":597544.37,
 "x":392.19,
 "y":304.0,
 "velocity":24.042,
 "player_id":259,
 "player_type":"vehicle.carlamotors.carlacola",
 "attributes":{
  "number_of_wheels":"4",
  "sticky_control":"true",
  "object_type":"",
  "color":"255,68,0",
  "role_name":"hero"
            },
    "weather_type": "WeatherParameters(cloudiness=30.0, cloudiness=30.0, precipitation=40.0, precipitation_deposits=40.0, wind_intensity=30.0, sun_azimuth_angle=250.0, sun_altitude_angle=20.0, fog_density=15.0, fog_distance=50.0, fog_falloff=0.9, wetness=80.0)"
}
\end{lstlisting}


Listing \ref{fig:JsonEx} shows one of the frames labeled in the \emph{json} file, where the aforementioned keys can be observed. In addition, the time elapsed between two consecutive images, the \emph{X} and \emph{Y} positions of the vehicle, and the vehicle's speed can be verified alternatively with this data.

\begin{figure*}[t]
    \centering
    \includegraphics[width=0.9\linewidth]{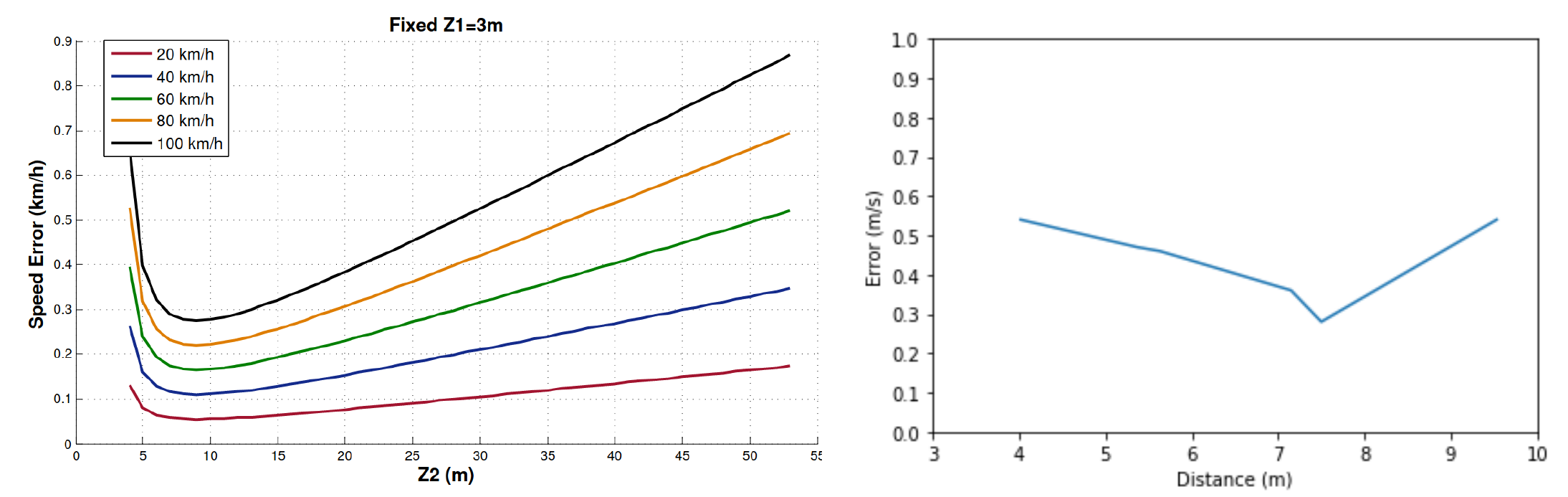}
    \caption{Left: theoretical speed estimation error with respect to the distance of the measurement section. From \cite{Llorca2016}. Right: obtained results of the speed estimation error with respect to the distance of the measurement section.}
    \label{fig:DErelation}
\end{figure*}

\subsection{3D-CNN model}

On this occasion, although the VGG16-GRU network used previously \cite{Hernandez2021} obtained highly successful results, it was decided to continue with the use of the 3D-CNN network, since it showed better performance.

Hence, in order to compare the different results obtained in each training performed, it has been decided not to modify any of the parameters or hyper-parameters of the neural network.

First of all, it has been decided to perform an individual training for each of the mentioned datasets, in this way it has been possible to check which of the configurations used is better for the proposed system.

After this, it has been decided to perform a training on the complete dataset, with all the sequences of each of the individual datasets, in order to check the generalization capacity of the network.

Finally, this model trained with the complete set of images has been tested with each of the test sets of the different individual datasets. In order to perform this type of tests, it has been guaranteed that the holistic model has not seen any of the test images of the individual models in its training.

\subsection{Training Parameters}

In this case, the trained network has not been modified in terms of its parameters with respect to the previous work. This is due to the fact that in this way it is possible to compare in a more correct way the different training with different datasets.

These parameters have been the following: a learning rate of $3\times10^{-4}$, a batch size of 5, Adam has been used as the optimizer, and the MSE has been used as the loss function. The network has been trained using the early-stopping technique, with a patience of 7, and the velocity data has been normalized between -1 and 1. In this way, the new training can be perfectly compared with the one performed in \cite{Hernandez2021}.

The complete dataset of 3660 sequences has been divided into 2196 training sequences (60\%), 732 validation sequences (20\%) and 732 test sequences (20\%). Furthermore, it has been guaranteed that the training images correspond to the training images of each of the 6 individual datasets, as well as the validation and test images, preventing the network with the holistic dataset from seeing test images from any of the individual datasets during training when obtaining crossed results.

\begin{figure*}[ht]
    \centering
    \includegraphics[width=0.8\linewidth]{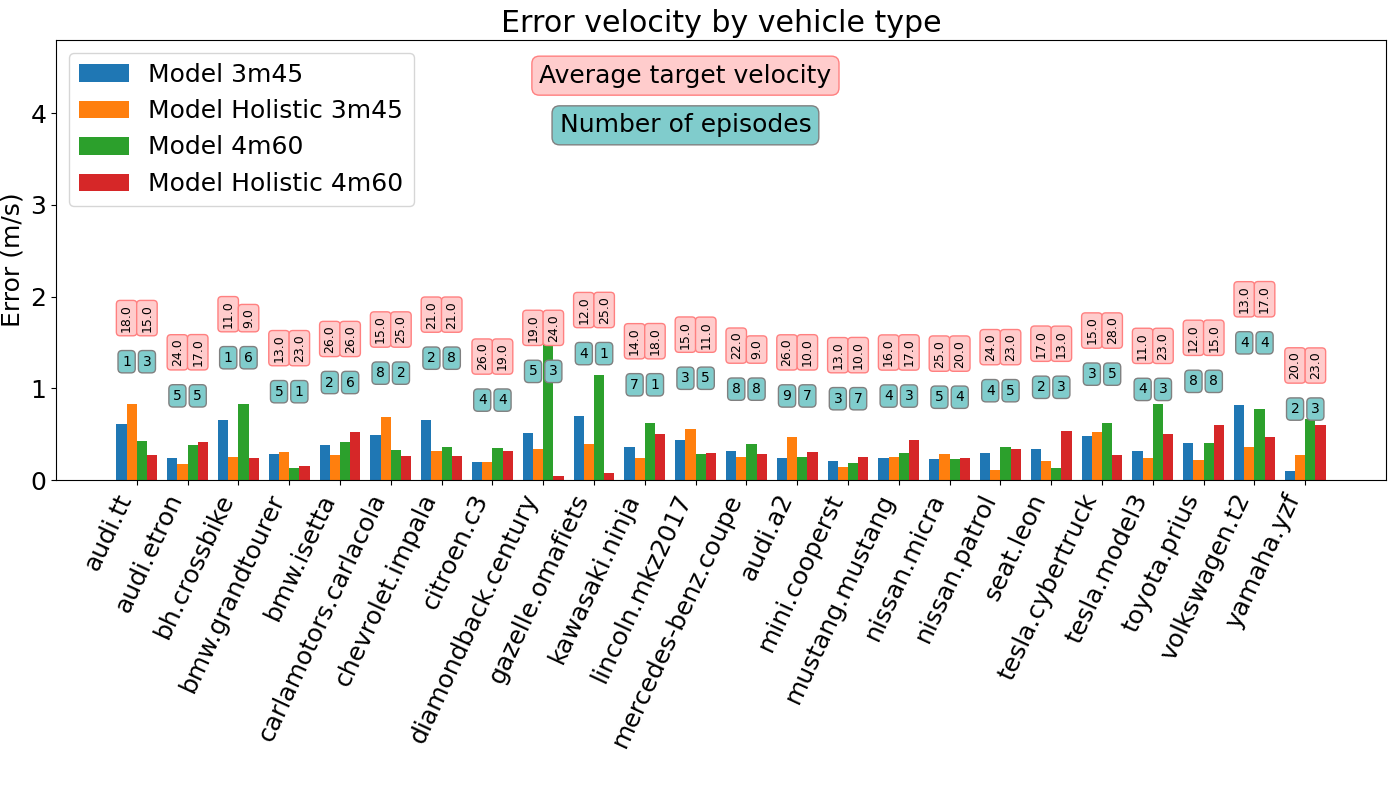}
    \caption{Error by Car Type. 3m 45º - 4m 60º}
    \label{fig:ErrorCarType3m45-4m60}
\end{figure*}

\begin{figure}[ht]
    \centering
    \includegraphics[width=1\linewidth]{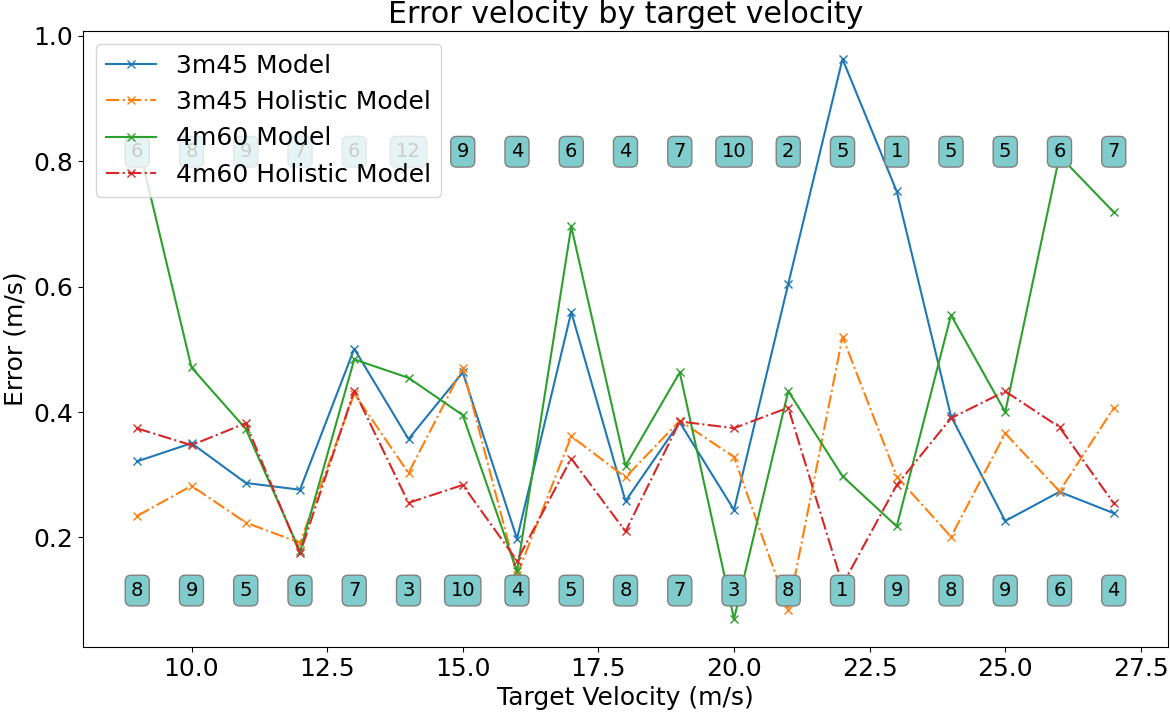}
    \caption{Error by Target Speed.  3m 45º - 4m 60º}
    \label{fig:ErrorTargetSpeed3m45-4m60}
    \vspace{0mm}
\end{figure}

\section{\uppercase{Experimental Evaluation}}
\label{sec:results}


The results obtained from the different datasets are generally very satisfactory. Table \ref{tab:ErrBySet} shows the average error obtained in the test set
 for the individual an holistic datasets.

\begin{table}
    \centering
    \begin{tabular}{c|c|c}
        \hline
        \textbf{Dataset} & \multicolumn{2}{c}{\textbf{Error [m/s]}} \\
    
        & Individual Mod.& Holistic Mod.\\ 
        
        \hline
        Only 3m 45º & 0.36 & \textbf{0.31} \\
        Only 3m 50º & 0.46 & \textbf{0.45} \\
        Only 3m 60º & 0.54 & \textbf{0.47} \\
        Only 4m 45º & 0.54 & \textbf{0.29} \\
        Only 4m 50º & \textbf{0.28} & 0.32 \\
        Only 4m 60º & 0.47 & \textbf{0.35} \\
        All dataset & - & \textbf{0.35} \\  \hline
    \end{tabular}
    \caption{Error by model type.}
    \label{tab:ErrBySet}
\end{table}

In the same way, it has been decided to use the holistic model trained with the total set of 3660 sequences, to test individually on each of the individual sets, obtaining the results shown in Table \ref{tab:ErrBySet}. To perform this it is important to remember that the images entered from each dataset belong to its own test set, so the holistic network has never seen those images.


As can be seen, by using the holistic model, the performance of the network has been improved in almost all cases. This is explained by the fact that this model has been trained using six times more sequences, and therefore has had more data to improve its performance. From these extra sequences that go into the training model, it can be deduced that with small variations the network is able to learn and generalise, avoiding in certain cases the use of a pre-trained model for each camera configuration.

In addition, it is possible to observe a behaviour that would be expected in the evolution of the error as the "mini-stretch" of road to be studied increases. The speed detection error reaches a minimum and, if we continue to reduce the stretch of road to be studied, the error increases again significantly, as explained in the work \cite{Llorca2016}. This can be seen in Fig. \ref{fig:DErelation}, where a minimum error can be observed with stretches of around 7 metres length.

As can be seen in the figure \ref{fig:ErrorCarType3m45-4m60}, in the case of the dataset with a camera located at a height of 3m and 45$^{\circ}$ pitch, the results of the holistic model slightly improve or equal the results obtained with the individual model of 3m and 45$^{\circ}$. This can be explained by the fact that this configuration is close to a stretch length captured by the camera of about 7m distance, which, as can be seen in the Fig. \ref{fig:DErelation}, is close to the point of minimum error, so the margin of improvement for this configuration would be very small. Likewise, the graph shows that the vehicles that show an improvement in error are precisely those in which bicycles have been simulated, vehicles whose behaviour was more difficult to predict by the individual model, and therefore, with a greater possibility of error reduction.

Similarly, for the configuration of the camera at a height of 4m and a pitch of 60$^{\circ}$, there is a considerable improvement in most cases, as this model had worse results when using its own model. Moreover, this can also be seen in Table \ref{tab:ErrBySet}, where we can see that when testing the dataset with the model that only knows the own images, the average error is 0.46m/s, while when using the holistic model, this error has been reduced to 0.33m/s.

On the other hand, in Figs. \ref{fig:ErrorCarType3m45-4m60} and \ref{fig:ErrorTargetSpeed3m45-4m60}, it can also be seen that while the 3m-45º dataset shows little improvement when using the holistic model, the 4m-60º dataset does improve the error considerably, especially at high speeds.

\begin{figure*}[ht]
    \centering
    \includegraphics[width=0.8\linewidth]{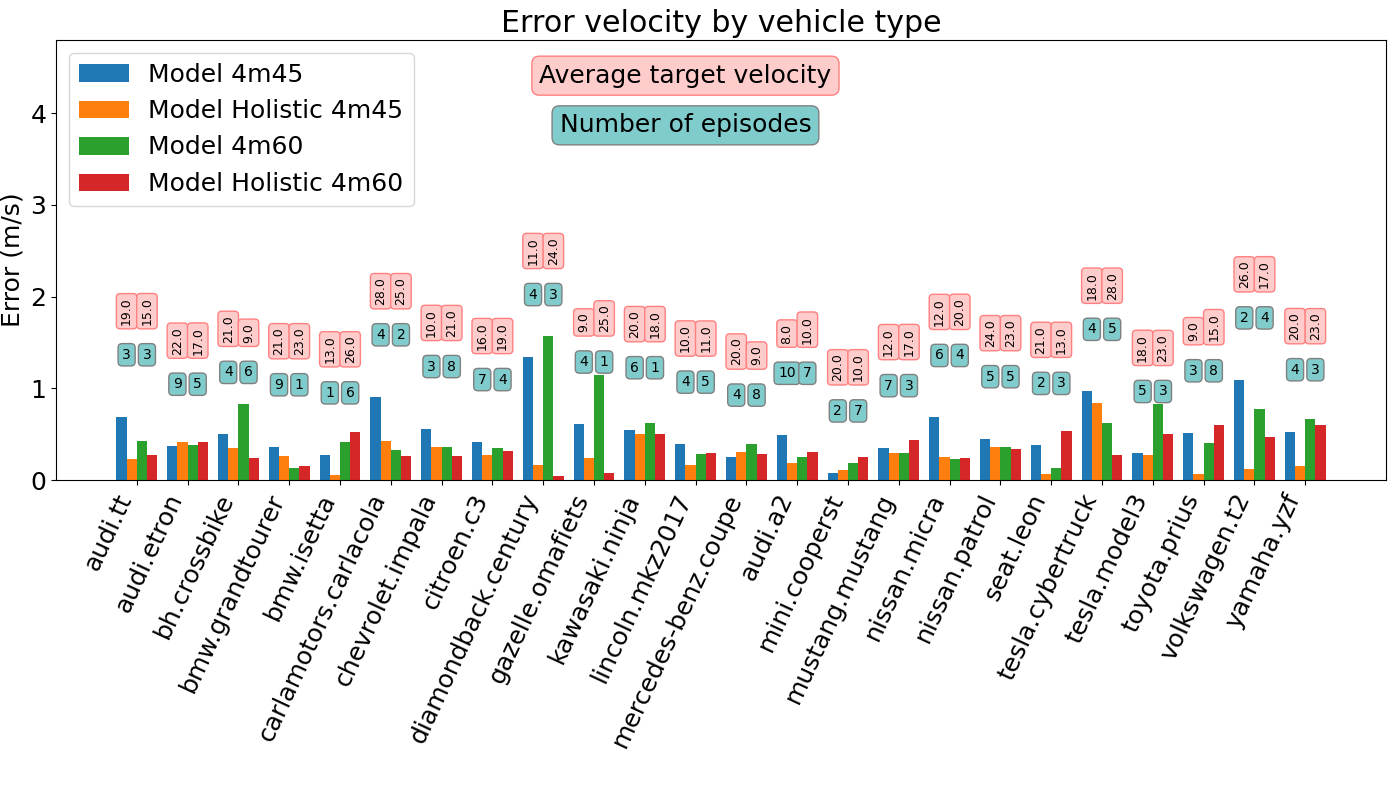}
    \caption{Error by Car Type. 4m 45º - 4m 60º}
    \label{fig:ErrorCarType4m45-4m60}
\end{figure*}

\begin{figure}[ht]
    \centering
    \includegraphics[width=1\linewidth]{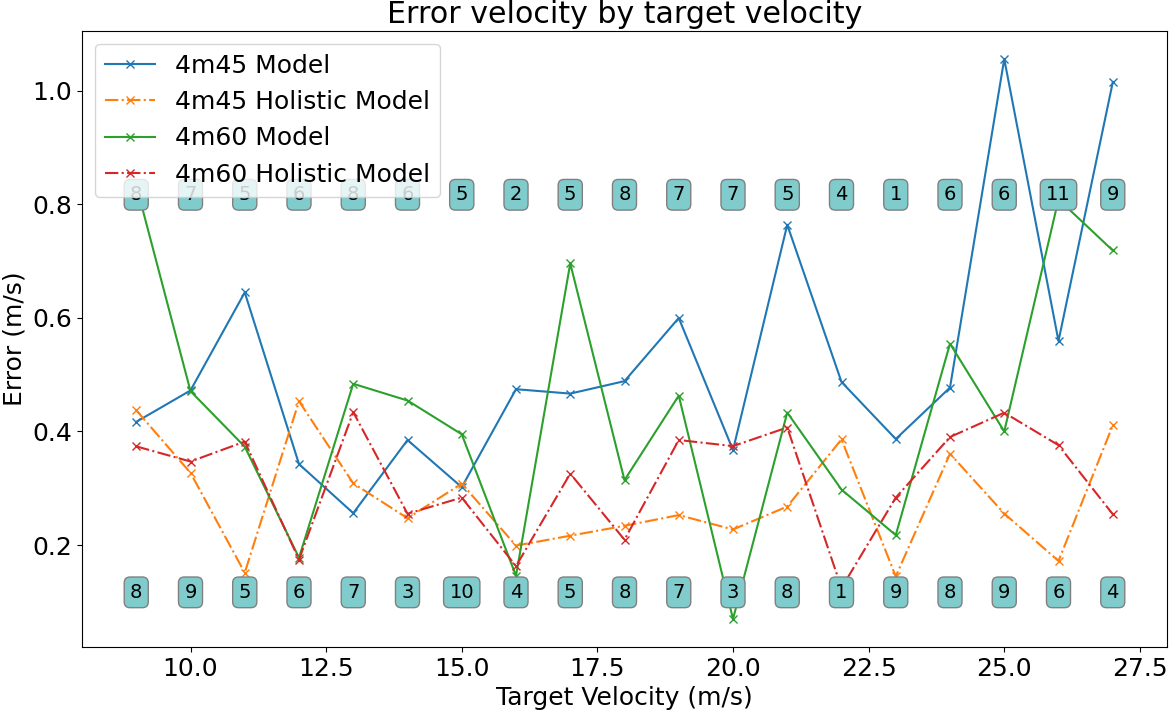}
    \caption{Error by Target Speed. 4m 45º - 4m 60º}
    \label{fig:ErrorTargetSpeed4m45-4m60}
\end{figure}

In Figs. \ref{fig:ErrorCarType4m45-4m60} and \ref{fig:ErrorTargetSpeed4m45-4m60} the same behaviour can be observed again. The holistic model improves in almost all cases the individual model, and this can be explained in the same way as in the previous case, since for both configurations of 4m 45º and 4m 60º they record a lane stretch of approximately 9m and 5m respectively, which is far from the minimum error zone.

Finally, recordings were made in a real environment at speeds of 30km/h, 50km/s and 70 km/h, using as ground truth the cruise control of a personal vehicle and a LiDAR system positioned perpendicular to the plane of the lane. Using the holistic model trained in simulation, individual errors were obtained for each of the speeds of 4.12 m/s, 1.48 m/s and 0.93 m/s respectively.

\begin{figure*}[ht]
    \centering
    \includegraphics[width=1\linewidth]{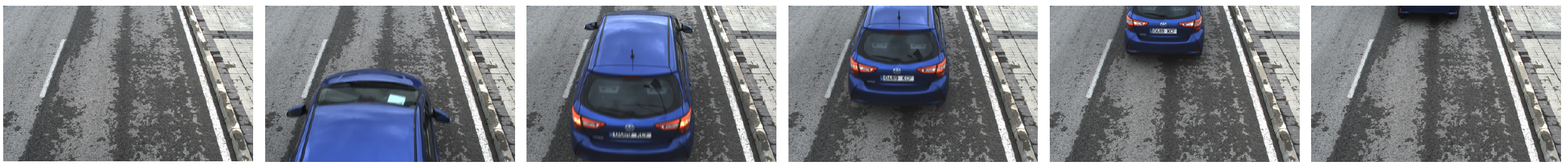}
    \caption{Real environment test. Sequence with a vehicle at 70km/h}
    \label{fig:RealEnv70}
\end{figure*}

These results can be considered satisfactory because it is a first preliminary test, and the recordings still have errors to correct, such as shutter and frame rate, to match the images captured in simulation.

An example of a vehicle driving at 70km/h is shown in the Fig. \ref{fig:RealEnv70}, where one of the problems identified when testing the recording, such as blur, can be observed. The images taken have a resolution of 1920x1208 and have been recorded at 20 fps. To carry out these tests, the camera was placed on a road with two lanes in each direction of travel, at a height of 3.5 metres and centred on the lane to be recorded. At the time of recording, the day was cloudy with no rain.

\section{\uppercase{Conclusions and Future works}}

We can prove that a holistic model, which is trained with a wide set of camera positions, can work for all of them, without the need to train a specific model for each of the positions. Moreover, as has been shown with real environment recordings, this model, which has never seen this kind of images before, is able to perform correctly.

These results indicate that it is possible to train a model with an even larger set of camera positions, and use it as a generic model for new and unknown positions for the network, obtaining sufficiently accurate results.

The results obtained indicate that, if a model were trained with a sufficiently large dataset, it could be capable of detecting the speed at which a vehicle is travelling with greater accuracy than current radar-type systems. Furthermore, it has also been shown that this is also valid for other types of vehicles such as motorbikes, since a considerable improvement in error has been seen in this type of vehicle with respect to the results obtained using only individual models. Finally, a global model for any camera position would avoid the work of having to train an individual model every time a new system is placed in the real environment.

As future work, the plan is to take recordings of longer duration, and in real traffic situations in an urban environment, as well as to extend the simulation training set, with new camera positions and different traffic directions, and to test it with different datasets in which there are recordings in real environment in different locations. In addition, the system can be tested on a multi-lane road, trying to detect the speed of vehicles in each lane individually, using computer vision techniques.After this, we could try to implement a real time system on roads, obtaining data such as the system's computation speed, and also, being able to be compared with other types of radars, whose errors in the national territory, according to the atuorities, oscillates between 5 and 7 km/h for speeds below 100km/h.

\section*{\uppercase{Acknowledgements}}
This work has been funded by research grant CLM18-PIC051 (Community Region of Castilla la Mancha), and partially funded by research grants S2018/EMT-4362 SEGVAUTO 4.0-CM (Community Region of Madrid) and PID2020-114924RB-I00 (Spanish Ministry Science Innovation)

\bibliographystyle{apalike}
{\small
\bibliography{ick3-22}}

\begin{thebibliography}{}

\bibitem[Barros and Oliveira, 2021]{Barros2021}
Barros, J. and Oliveira, L. (2021).
\newblock Deep speed estimation from synthetic and monocular data.
\newblock In {\em 2021 IEEE Intelligent Vehicles Symposium (IV)}, pages
  668--673.

\bibitem[Biparva et~al., 2022]{Biparva2022}
Biparva, M., Llorca, D.~F., Izquierdo, R., and Tsotsos, J.~K. (2022).
\newblock Video action recognition for lane-change classification and
  prediction of surrounding vehicles.
\newblock {\em IEEE Transactions on Intelligent Vehicles}.

\bibitem[Dong et~al., 2019]{Dong2019}
Dong, H., Wen, M., and Yang, Z. (2019).
\newblock Vehicle speed estimation based on 3d convnets and non-local blocks.
\newblock {\em Future Internet}, 11.

\bibitem[Dosovitskiy et~al., 2017]{carla2017}
Dosovitskiy, A., Ros, G., Codevilla, F., Lopez, A., and Koltun, V. (2017).
\newblock {CARLA}: {An} open urban driving simulator.
\newblock In {\em Proceedings of the 1st Annual Conference on Robot Learning},
  pages 1--16.

\bibitem[Elvik, 2011]{Elvik2011}
Elvik, R. (2011).
\newblock Developing an accident modification function for speed enforcement.
\newblock {\em Safety Science}, 49:920--925.

\bibitem[{ERSO, EC}, 2018]{ERSO2018}
{ERSO, EC} (2018).
\newblock Speed enforcement 2018.
\newblock {\em European Road Safety Observatory}.

\bibitem[Fern\'{a}ndez-Llorca et~al., 2021]{Llorca2021}
Fern\'{a}ndez-Llorca, D., Hern\'{a}ndez-Mart\'{i}nez, A., and Garc\'{i}a-Daza,
  I. (2021).
\newblock Vision-based vehicle speed estimation: A survey.
\newblock {\em IET Intelligent Transport Systems}, 15(8):987--1005.

\bibitem[Lee et~al., 2019]{Lee2019}
Lee, J., Roh, S., Shin, J., and et~al. (2019).
\newblock Image-based learning to measure the space mean speed on a stretch of
  road without the need to tag images with labels.
\newblock {\em Sensors}, 19.

\bibitem[{Llorca} et~al., 2016]{Llorca2016}
{Llorca}, D.~F., {Salinas}, C., {Jiménez}, M., {Parra}, I., {Morcillo}, A.~G.,
  {Izquierdo}, R., {Lorenzo}, J., and {Sotelo}, M.~A. (2016).
\newblock Two-camera based accurate vehicle speed measurement using average
  speed at a fixed point.
\newblock In {\em 2016 IEEE 19th International Conference on Intelligent
  Transportation Systems (ITSC)}, pages 2533--2538.

\bibitem[Luvizon et~al., 2017]{Luvizon2017}
Luvizon, D.~C., Nassu, B.~T., and Minetto, R. (2017).
\newblock A video-based system for vehicle speed measurement in urban roadways.
\newblock {\em IEEE Transactions on Intelligent Transportation Systems},
  18:1393--1404.

\bibitem[Madhan et~al., 2020]{Madhan2020}
Madhan, E.~S., Neelakandan, S., and Annamalai, R. (2020).
\newblock A novel approach for vehicle type classification and speed prediction
  using deep learning.
\newblock {\em Journal Comp. Theor. Nano.}, 17:2237--2242.

\bibitem[Martinez et~al., 2021]{Hernandez2021}
Martinez, A.~H., Díaz, J.~L., Daza, I.~G., and Llorca, D.~F. (2021).
\newblock Data-driven vehicle speed detection from synthetic driving simulator
  images.
\newblock In {\em 2021 IEEE International Intelligent Transportation Systems
  Conference (ITSC)}, pages 2617--2622.

\bibitem[Parimi and Jiang, 2021]{Parimi2021}
Parimi, A. and Jiang, Z. (2021).
\newblock Dynamic speed estimation of moving objects from camera data.
\newblock In {\em NAECON 2021 - IEEE National Aerospace and Electronics
  Conference}, pages 307--316.

\bibitem[Revaud and Humenberger, 2021]{Revaud2021}
Revaud, J. and Humenberger, M. (2021).
\newblock Robust automatic monocular vehicle speed estimation for traffic
  surveillance.
\newblock In {\em 2021 IEEE/CVF International Conference on Computer Vision
  (ICCV)}, pages 4531--4541.

\bibitem[Rodríguez-Rangel et~al., 2022]{RR2022}
Rodríguez-Rangel, H., Morales-Rosales, L.~A., Imperial-Rojo, R., Roman-Garay,
  M.~A., Peralta-Peñuñuri, G.~E., and Lobato-Báez, M. (2022).
\newblock Analysis of statistical and artificial intelligence algorithms for
  real-time speed estimation based on vehicle detection with yolo.
\newblock {\em Applied Sciences}, 12(6).

\bibitem[Sochor et~al., 2017]{Sochor2017}
Sochor, J., Juránek, R., and Herout, A. (2017).
\newblock Traffic surveillance camera calibration by 3d model bounding box
  alignment for accurate vehicle speed measurement.
\newblock {\em Computer Vision and Image Understanding}, 161:87--98.

\bibitem[Wilson et~al., 2010]{Wilson2010}
Wilson, C., Willis, C., Hendrikz, J.~K., and et~al. (2010).
\newblock Speed cameras for the prevention of road traffic injuries and deaths.
\newblock {\em Cochrane Database of Systematic Reviews}, 11.

\end{thebibliography}



\end{document}